\PassOptionsToPackage{obeyspaces}{url}
\documentclass[sigconf,screen]{acmart}
\usepackage[acm,subfig]{definition}

\newcommand{\themodel}{\textsf{ALDEN}\xspace}

\copyrightyear{2021} 
\acmYear{2021} 
\setcopyright{acmlicensed}
\acmConference[CIKM '21]{Proceedings of the 30th ACM International Conference on Information and Knowledge Management}{November 1--5, 2021}{Virtual Event, QLD, Australia}
\acmBooktitle{Proceedings of the 30th ACM International Conference on Information and Knowledge Management (CIKM '21), November 1--5, 2021, Virtual Event, QLD, Australia}
\acmPrice{15.00}
\acmDOI{10.1145/3459637.3482080}
\acmISBN{978-1-4503-8446-9/21/11}

\begin{document}
\fancyhead{}

\title{Deep Active Learning for Text Classification \mbox{with Diverse Interpretations}}

\author[Qiang Liu, Yanqiao Zhu, Zhaocheng Liu, Yufeng Zhang, and Shu Wu]{Qiang Liu$^{1,2}$, Yanqiao Zhu$^{1,2}$, Zhaocheng Liu$^{3}$, Yufeng Zhang$^{1}$, and Shu Wu$^{1,2,}$*}

\authornotetext{To whom correspondence should be addressed.}

\affiliation{
    \institution{$^1$Center for Research on Intelligent Perception and Computing, Institute of Automation, Chinese Academy of Sciences}
    \institution{$^2$School of Artificial Intelligence, University of Chinese Academy of Sciences \quad $^3$RealAI}
    \country{}
}

\email{{qiang.liu, shu.wu}@nlpr.ia.ac.cn, {yanqiao.zhu, yufeng.zhang}@cripac.ia.ac.cn, lio.h.zen@gmail.com}

\def\authors{Qiang Liu, Yanqiao Zhu, Zhaocheng Liu, Yufeng Zhang, and Shu Wu}

\begin{abstract}
Recently, Deep Neural Networks (DNNs) have made remarkable progress for text classification, which, however, still require a large number of labeled data.
To train high-performing models with the minimal annotation cost, active learning is proposed to select and label the most informative samples, yet it is still challenging to measure informativeness of samples used in DNNs.
In this paper, inspired by piece-wise linear interpretability of DNNs, we propose a novel Active Learning with DivErse iNterpretations (\themodel) approach.
With local interpretations in DNNs, \themodel identifies linearly separable regions of samples.
Then, it selects samples according to their diversity of local interpretations and queries their labels.
To tackle the text classification problem, we choose the word with the most diverse interpretations to represent the whole sentence.
Extensive experiments demonstrate that \themodel consistently outperforms several state-of-the-art deep active learning methods.
\end{abstract}


\maketitle

\section{Introduction}

In recent years, Deep Neural Networks (DNNs) have achieved the state-of-the-art supervised performance in numerous research tasks.
Among them, a typical task in natural language processing is text classification, where deep models such as Convolutional Neural Networks (CNNs) \cite{Kim:2014Convolutional} and Recurrent Neural Networks (RNNs) \cite{Wang:2019Convolutional} are often adopted.
However, such deep models require a large number of labeled samples, which are expensive and labor-consuming to obtain in real-world applications.
Fortunately, active learning, which aims to identify and label the most informative samples from a pool of unlabeled data to train deep models with limited labels, is a promising approach to relieve this problem \cite{Wang:2015Querying,Zhang:2017Active,Ash:2020Deep,Ein:2020Active}.

Existing works on active learning mainly select samples based on uncertainty and diversity.
Taking Expected Gradient Length (EGL) \cite{Huang:2016Active} as an example, it computes the sample uncertainty as the norms of gradients of losses with respect to the model parameters.
Following EGL, EGL-Word \cite{Zhang:2017Active} selects the word with the largest EGL among all samples to query its label so as to maximize the model performance for text classification.
In addition, Bayesian Active Learning by Disagreement (BALD) \cite{Gal:2017Deep} measures the uncertainty according to the probabilistic distribution of the model output via Bayesian inference, where an approximation by dropout is usually incorporated \cite{Gal:2016Dropout}.
On the other hand, to measure the diversity of samples, some works define the active learning task as a CORESET problem \cite{Sener:2018Active} and uses the embedding of the last layer in deep models as the representation of samples.
There are also attempts to trade off between uncertainty and diversity \cite{Huang:2014Active,Wang:2015Querying}.
For example, Batch Active learning by Diverse Gradient Embeddings (BADGE) \cite{Ash:2020Deep} can be viewed as a combination of EGL and CORESET.
Meanwhile, there are empirical experiments to evaluate above approaches on text classification \cite{Siddhant:2018Deep,Prabhu:2019Sampling,Yan:2020Active,Ein:2020Active}.

\begin{figure*}
	\centering
	\subfloat[Data distribution]{
		\label{fig:distribution}
		\includegraphics[width=0.24\linewidth]{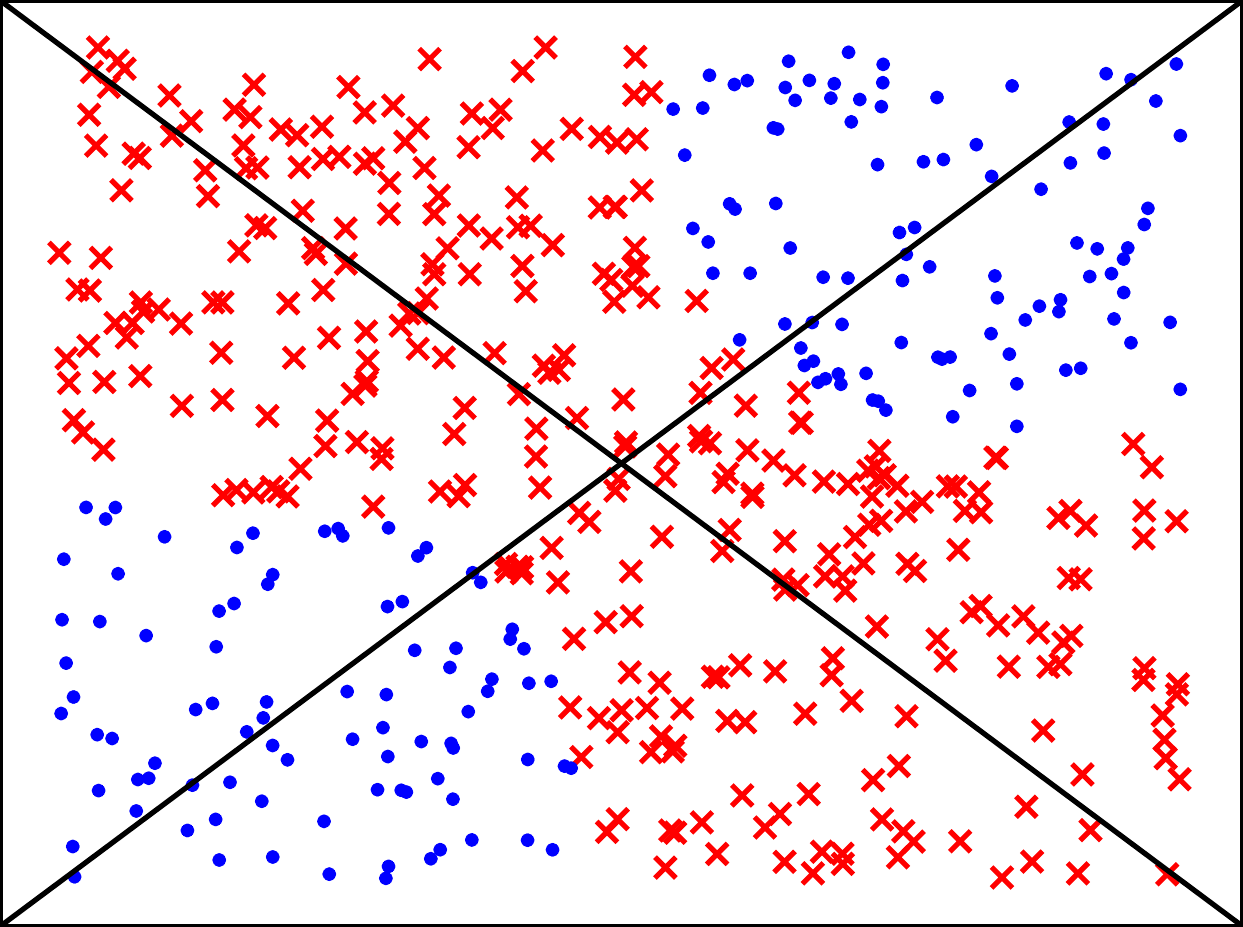}
	}
	\subfloat[Clustering with CORESET \cite{Sener:2018Active}]{
		\label{fig:clustering1}
		\includegraphics[width=0.24\linewidth]{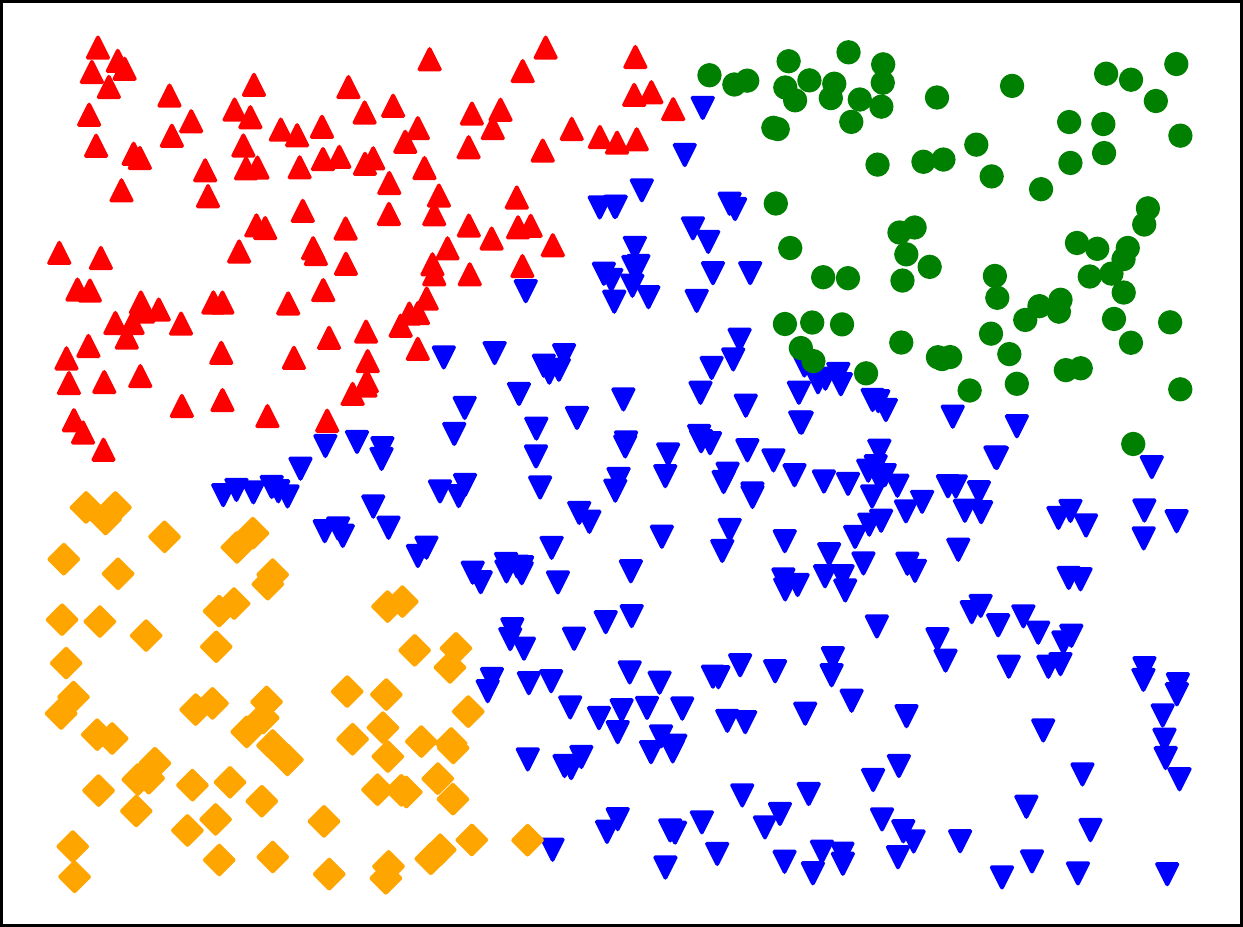}
	}
	\subfloat[Clustering with BADGE \cite{Ash:2020Deep}]{
		\label{fig:clustering2}
		\includegraphics[width=0.24\linewidth]{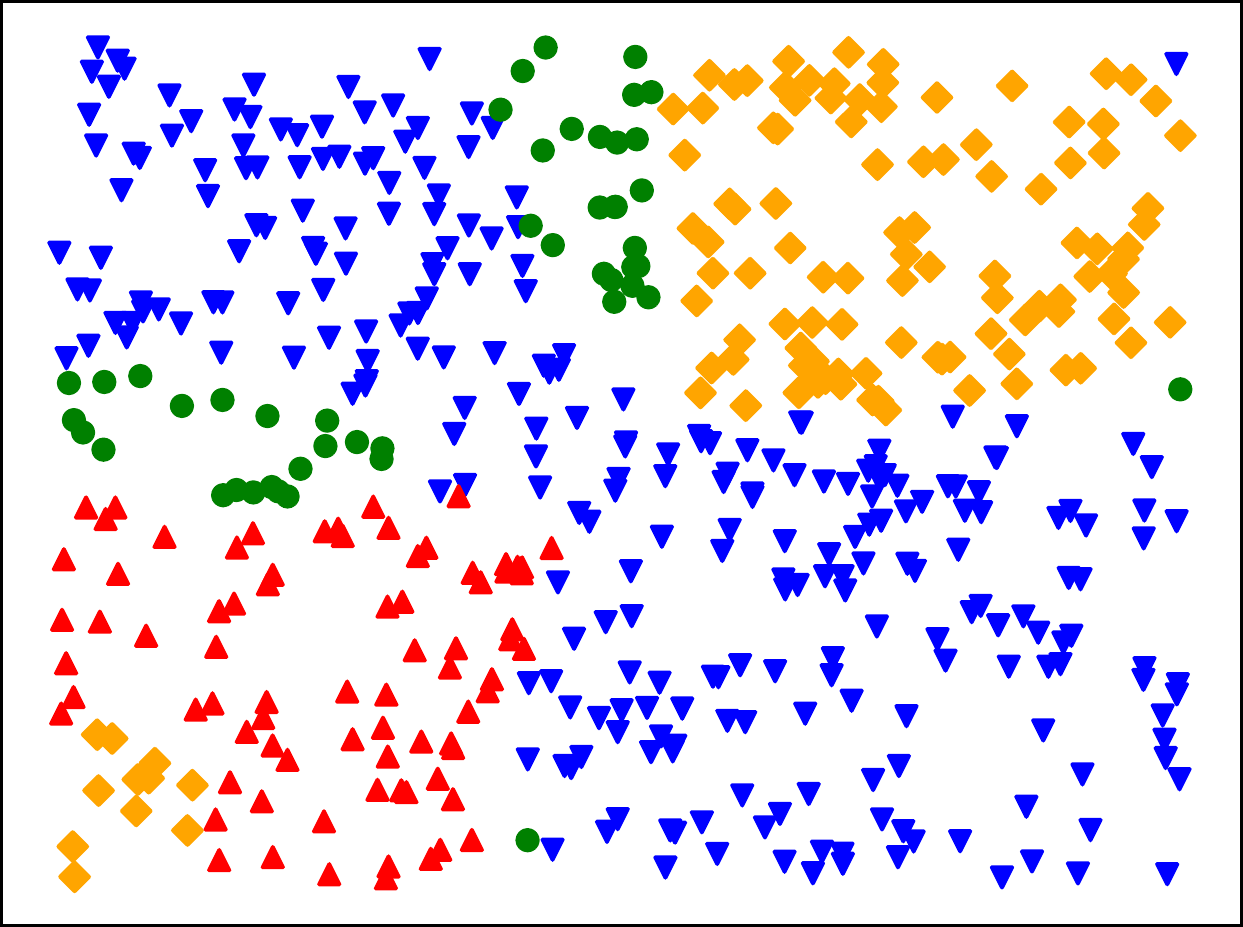}
	}
	\subfloat[Clustering with local interpretations]{
		\label{fig:clustering3}
		\includegraphics[width=0.24\linewidth]{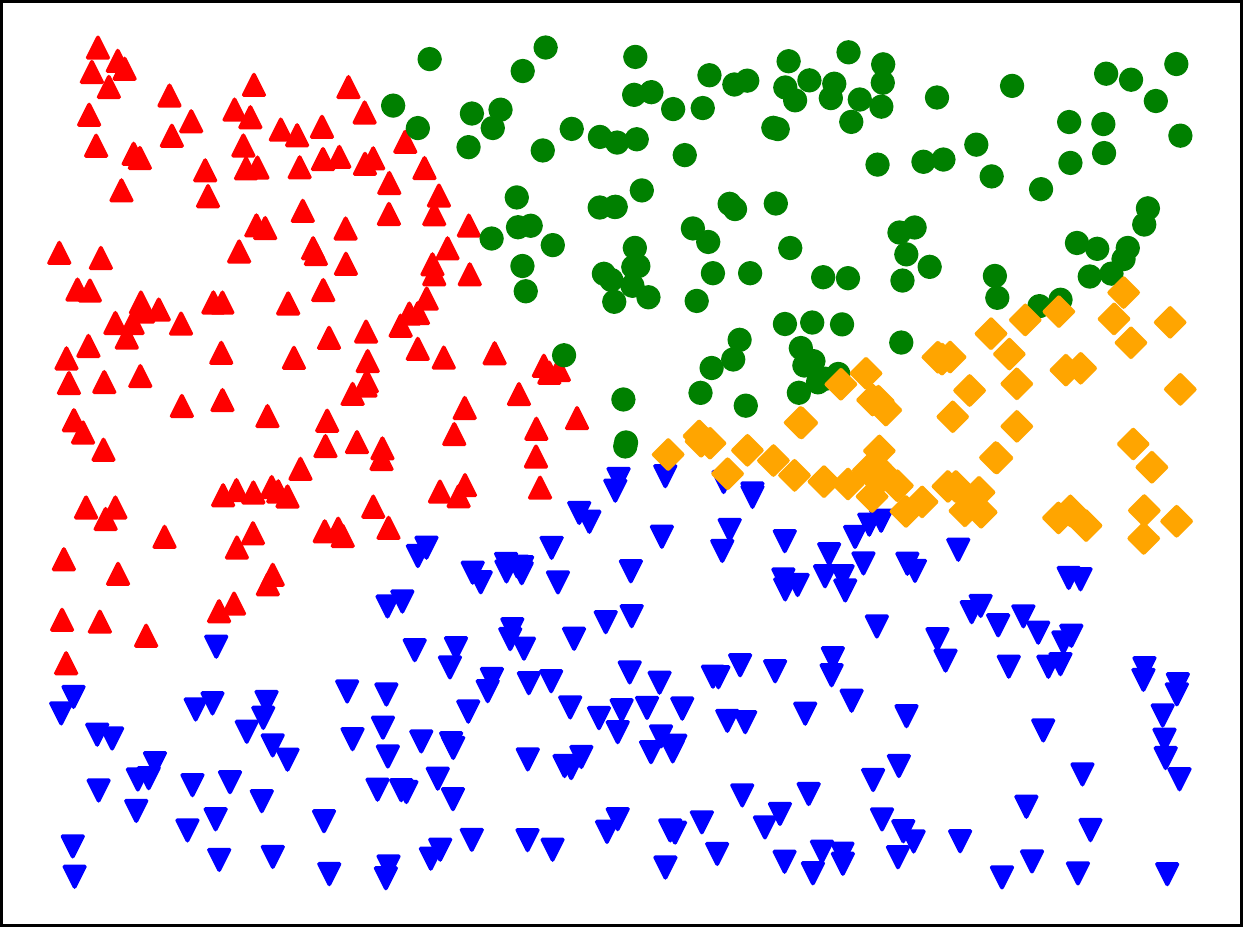}
	}
	\caption{Illustrating local interpretations in DNNs. We artificially generate a series of data samples that could be roughly divided into four linearly separable regions (shown in four triangle areas). We perform \(k\)-Means clustering on the example data, where the representations of samples are from CORESET \cite{Sener:2018Active} and BADGE \cite{Ash:2020Deep}, as well as the local interpretations in DNNs computed using Eq. (\ref{equation:local_interpretation}). The clusters are shown in four different colors. It is seen that only with interpretations we are able to correctly identify the four linearly separable regions.}
	\label{fig:cluster}
\end{figure*}

Recently, the interpretability of DNNs has received increasingly attention, among which most works focus on local piece-wise interpretability \cite{Ribeiro:2016Why,Chu:2018Exact}.
To be specific, previous works \cite{Mont:2014On,Harvey:2017Nearly,Chu:2018Exact} investigate the local interpretability of DNNs and show that a deep model with piece-wise linear activations, e.g., Maxout \cite{Goodfellow:2013Maxout} and the family of ReLU \cite{Nair:2010Rectified,Glorot:2011Deep}, can be regarded as a set of numerous local linear classifiers.
The linear separable regions corresponding to these linear classifiers can be determined by the local piece-wise interpretations in DNNs that are calculated via gradient backpropagation \cite{Li:2016Visualizing,Selvaraju:2020Grad,Smilkov:2017SmoothGrad,Yuan:2019Interpreting} or feature perturbation \cite{Fong:2017Interpretable,Guan:2019Towards}.
In other words, samples used in a DNN could be divided into numerous linearly separable regions according to their local interpretations and samples in the same linearly separable region are classified by the same local linear classifier \cite{Chu:2018Exact}.
Therefore, fitting a DNN model is roughly equivalent to fitting all the linear classifiers in different linearly separable regions.
Inspired by this, we propose to actively select samples in different linearly separable regions with the maximally diverse local interpretations, so that linear classifiers in different linearly separable regions can be all well trained.

In this paper, we propose a novel Active Learning with DivErse iNterpretations (\themodel) approach for text classification.
In our proposed approach, we first calculate the local interpretations in DNN for each sample as the gradient backpropagated from the final predictions to the input features \cite{Li:2016Visualizing,Selvaraju:2020Grad}.
Then, we use the most diverse interpretation of words in a sample to measure its diverseness. Accordingly, we select unlabeled samples with the maximally diverse interpretations for labeling and retrain the model with these labeled samples.
We conduct experiments on two text classification datasets, with two representative deep classifiers: CNN \cite{Kim:2014Convolutional} and Bi-directional Long Short-Term Memory (BiLSTM).
Extensive experimental results show that \themodel can constantly outperform state-of-the-art deep active learning approaches.


\section{Local Interpretations in Deep Neural Networks} \label{sec:2}

Recently, extensive works have been conducted to study local piece-wise interpretability of DNNs, which can be computed using the gradient backpropagation from the predictions to the input features \cite{Selvaraju:2020Grad,Li:2016Visualizing,Smilkov:2017SmoothGrad,Yuan:2019Interpreting,Liu:2021DNN2LR}.
To be specific, we first train a deep model and obtain the prediction ${\hat y}_i$ given input features $x_i$ of a specific sample. Then, we can calculate local interpretations as
\begin{equation} \label{equation:local_interpretation}
I_{i}  = \frac{{\partial \, {\hat y}_i }}{{\partial \, x_{i} }}.
\end{equation}
As in \citet{Li:2016Visualizing}, local interpretations could be formulated by
\begin{equation} \label{equation:local_interpretation_app}
{\hat{y}}_i  \approx  {I_i \, x_i^\top + b},
\end{equation}
where $b$ is the bias term.
As mentioned in previous works \cite{Mont:2014On,Ribeiro:2016Why,Chu:2018Exact}, a DNN model with piece-wise linear activation functions (such as Maxout and ReLU \cite{Goodfellow:2013Maxout,Nair:2010Rectified,Glorot:2011Deep}) can be regarded as a combination of numbers of local linear classifiers, which are introduced by the local interpretations in DNN.
That is to say, local interpretations $I_i$ of sample $x_i$ as calculated in Eq. (\ref{equation:local_interpretation}) can be partitioned into several clusters and each of them corresponds to a specific local linear classifier.
With the local piece-wise interpretations in DNNs, samples can be divided into numerous linearly separable regions and samples in the same linearly separable region are classified by the same local linear classifier \cite{Chu:2018Exact}.
Therefore, fitting a DNN model means fitting all the linear classifiers in different linearly separable regions.
Accordingly, if we select samples according to diverse local interpretations, different linear classifiers in different linearly separable regions can be optimized in a more balanced way, so that the corresponding DNN model can be better trained.
Thus, we argue that adopting local interpretations in DNN could potentially benefit deep active learning.

To demonstrate that the local Interpretations in DNNs can help promote deep active learning, we present a concrete example as shown in Figure \label{fig:cluster}, where example data are drawn from a probability distribution $p( y_i  = 1 \mid x_i ) = \sigma ( x_{i,1} \cdot x_{i,2})$, where $x_{i,1}$ and $x_{i,2}$ are uniformly drawn from $[ -5.0,5.0 ]$, and $\sigma ( \cdot )$ is the sigmoid function.
The distribution of these artificially generated samples is shown in Figure \ref{fig:distribution}, which exhibits clear nonlinear characteristics. In addition, it is seen there are roughly four linearly separable regions, corresponding to the four triangle areas.
For these samples, we run \(k\)-Means clustering on the representations generated by CORESET \cite{Sener:2018Active} and BADGE \cite{Ash:2020Deep}, as well as local interpretations in a Multi-Layer Perception (MLP) model, all trained on the example data.
We set the number of clusters in \(k\)-Means to 4 and present the results in Figures \ref{fig:clustering1}, \ref{fig:clustering2}, and \ref{fig:clustering3} respectively.
We can observe that CORESET focuses on the original feature distribution and different classes, while BADGE pays more attention to the decision boundaries.
Clearly, we can only use local interpretations to distinguish the four linearly separable regions.
Therefore, with the help of local interpretations in DNNs, we are able to identify samples in different linearly separable regions. Inspired by this observation, we propose a deep active learning strategy to better fit all the linear classifiers corresponding to the DNN model.

\section{The Proposed \themodel Approach}

In this section, we introduce the \themodel approach for text classification in detail.

\subsection{Problem Formulation}

In this work, we apply pool-based active learning in the batch mode \cite{Settles:2009Active,Zhang:2017Active,Ein:2020Active,Yang:2018Active}.
Specifically, we have a small set of labeled samples $\mathcal L$ and a large set of unlabeled samples $\mathcal U$.
Sample $x_i \in \mathcal L$ is associated with label $y_i$, while sample $x_i \in \mathcal U$ has no labels.
The feature vector $x_i$ is denoted as $x_i = ( x_{i,1}, x_{i,2},...,x_{i,{|x_i|}} )$, where $x_{i,j}$ is a word in the sample.
With the labeled samples in $\mathcal L$, we can train a text classifier $f\left( x|\theta \right)$: $\mathcal X \to \mathcal Y$.
We need to develop an active learning strategy to select samples from $\mathcal U$ and add them to $\mathcal L$ for further training the classifier.
We set the label budget to \(K\) samples per iteration of sample selection and train the model for a total of \(N\) iterations.

\subsection{Approach Details}

Regarding active learning for text classification, similar to Eq. (\ref{equation:local_interpretation}), for a word in a specific sample used in a deep text classifier, we can compute its local interpretation as
\begin{equation} \label{equation:local_interpretation1}
{I_{i,j}} = \frac{\partial \,{\hat y}_i}{\partial \,x_{i,j}},
\end{equation}
where ${{\hat y}_i}$ is the prediction of sample $x_i$.
Equivalently, we can also calculate Eq. (\ref{equation:local_interpretation1}) using word embedding \(e_{i,j}\) of word \(x_{i,j}\):
\begin{equation} \label{equation:local_interpretation2}
{I_{i,j}} = \frac{\partial \,{\hat y}_i}{\partial\,e_{i,j}} e_{i,j}^ \top.
\end{equation}
Recall that the local interpretation of a word indicates its contribution to the final prediction; similar to Eq. (\ref{equation:local_interpretation_app}), the prediction can be approximated \cite{Li:2016Visualizing} as
\begin{equation} \label{equation:contribution}
{{\hat y}_i} \approx \sum\limits_{1 \le j \le \left| {{x_i}} \right|} {\frac{{\partial\,{{\hat y}_i}}}{{\partial\,{e_{i,j}}}}\mathop e\nolimits_{i,j}^ \top  }  + b.
\end{equation}

Consider that local interpretations (i.e. the contribution to the model predictions) of the same word may be different among different samples, due to the complex nonlinear feature interactions modeled by deep models \cite{Mont:2014On,Ribeiro:2016Why,Chu:2018Exact}.
As discussed in Section \ref{sec:2}, we need to select samples with the diverse local interpretations, so that linear classifiers in different linearly separable regions can be well optimized.
Meanwhile, since diverse interpretations indicate different decision regions in the deep model, samples with the maximally diverse interpretations can provide the most comprehensive information to learn diverse decision logics in the deep model.
For the task of text classification, as different samples consist of various numbers of words, we need to start with analyzing local interpretations of words in each sample.
In particular, we calculate the interpretation diversity of a word $x_{i,j}$ compared to the same word appeared in labeled samples
\begin{equation} \label{equation:diversity1}
D( {{\mathcal L},{x_{i,j}}} ) = \min_{\substack{
x_m \in {\mathcal L}\\
1 \le w \le \left| x_m \right|\\
x_{m,w} = x_{i,j}
}} \left\| I_{i,j} - I_{m,w} \right\|,
\end{equation}
which is similar to the distance calculation in the greedy \(k\)-Center algorithm \cite{Sener:2018Active}.
However, some words may not appear in the labeled samples, which makes it infeasible to directly calculate Eq. (\ref{equation:diversity1}).
As a remedy, we search for the most similar embedding of the word appearing in the labeled samples as the neighbor, which is formulated as
\begin{equation} \label{equation:neighbour}
N( {\mathcal L},x_{i,j} ) = \argmin_{\substack{
x_m \in {\mathcal L}\\
1 \le w \le \left| x_m \right|}}
\left\| e_{i,j} - e_{m,w} \right\|.
\end{equation}
Then, we can rewrite Eq. (\ref{equation:diversity1}) as
\begin{equation} \label{equation:diversity2}
D( {\mathcal L},x_{i,j} ) = \min_{\substack{
x_m \in {\mathcal L}\\
1 \le w \le \left| {{x_m}} \right|\\
x_{m,w} = N( L,x_{i,j} )}}
\left\| I_{i,j} - I_{m,w} \right\|.
\end{equation}

Considering that there are various numbers of words in sentences, which brings difficulties in directly using the local interpretations of all words in a sample. Therefore, we adopt a pooling strategy for active learning.
Recall that in EGL-Word \cite{Zhang:2017Active}, the word with the largest EGL is used to represent the whole sentence.
Aligning with EGL-Word, we also use the word with the maximally diverse interpretation to represent the whole sample for active learning.
Formally, for a sample $ x_i  \in {\mathcal U}$, we have
\begin{equation} \label{equation:diversity3}
D( {\mathcal L},x_i ) = \max_{1 \le j \le \left|x_i\right|} D( {\mathcal L},x_{i,j}).
\end{equation}
Based on the metric calculated using Eq. (\ref{equation:diversity3}), we can select the unlabeled sample that has the maximally diverse interpretation for labeling:
\begin{equation}
	x = \argmax_{ x_i  \in \mathcal U} D( \mathcal L, x_i).
\end{equation}
Since we are given a budget of $K$ in each iteration, we repeat the above process for $K$ times to select and label $K$ samples.
Algorithm \ref{alg:ALDI} summarizes the training procedure of the \themodel approach.

\begin{algorithm}
    \caption{The \themodel approach}
    \label{alg:ALDI}
    \KwData{Labeled samples $\mathcal L$, unlabeled samples $\mathcal U$, budget $K$ in each iteration, and the number of iterations $N$.}
    Train an initial model $f( x \mid \theta_0 )$ on $\mathcal L$\;
    \For{$n=1,2,...,N$}{
		\For{${x_i} \in {\mathcal L} \cup {\mathcal U}$}{
			Calculate prediction ${\hat y}_i  = f(  x_i \mid \theta_{n - 1}  )$\;
			\For{${1 \le j \le | x_i |}$}{
				Calculate the local interpretation $I_{i,j}$ according to Eq. (\ref{equation:local_interpretation2})\;
				}
			}
            \For{$k=1,2,...,K$}{
                \For{$x_i \in \mathcal U$}{
                    \For{${1 \le j \le | x_i |}$}{
						Find the neighbor $N( \mathcal L, x_{i,j} )$ of word $x_{i,j}$ according to Eq. (\ref{equation:neighbour})\;
						Compute the diversity $D( \mathcal L,x_{i,j} )$ of local interpretations of $x_{i,j}$ according to Eq. (\ref{equation:diversity2})\;
					}
					Compute the diversity $D( \mathcal L, x_i )$ of the local interpretation of $x_{i}$ according to Eq. (\ref{equation:diversity3})\;
				}
				Select and label the sample $x$ having the most diverse local interpretations\;
				$\mathcal L = \mathcal L \cup \{ x \}$\;
				$\mathcal U = \mathcal U \backslash x$\;
			}
			Train a new model $f( x \mid \theta_n )$ on $\mathcal L$\;
    }
    \Return{The final model $f( x \mid \theta_N )$}
\end{algorithm}

\section{Experiments}
In this section, we empirically evaluate our proposed \themodel approach on the task of text classification.

\begin{figure*}
\centering
\includegraphics[width=0.85\textwidth]{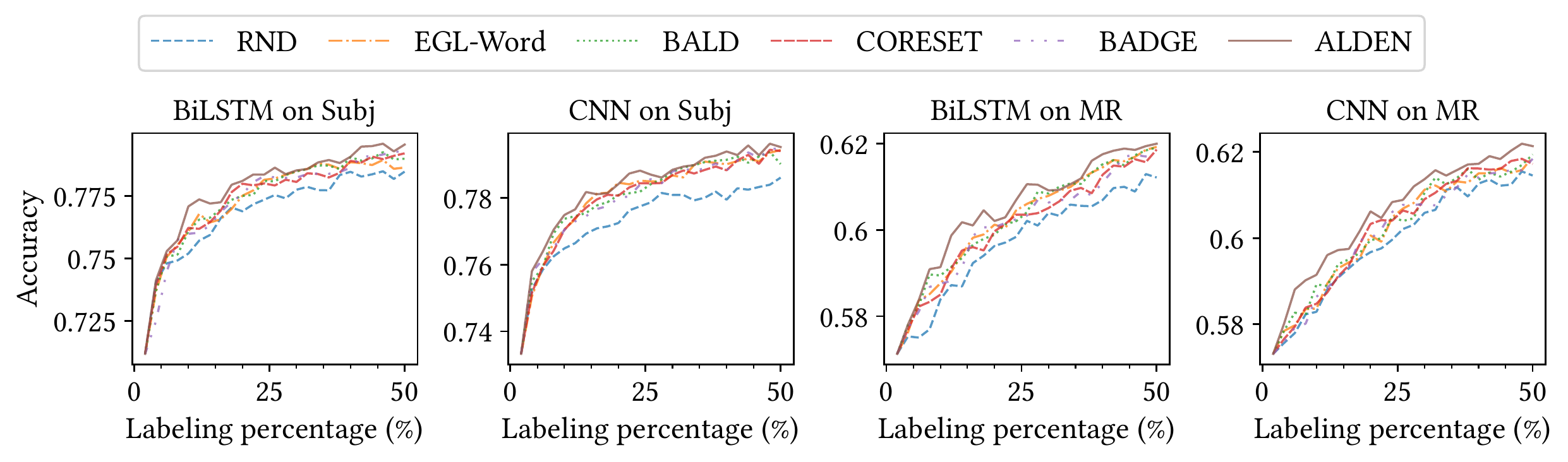}
\caption{Learning curves in terms of accuracy of compared approaches with various labeling rates of training samples.}
\label{fig:curve}
\end{figure*}

\begin{table}
  \centering
  \caption{Normalized area under curve scores of learning curves. The larger the values, the better the performances.}
    \begin{tabular}{ccccc}
    \toprule
    \multirow{2.5}{*}{Model} & \multicolumn{2}{c}{Subj} & \multicolumn{2}{c}{MR} \\
    \cmidrule(lr){2-3} \cmidrule(lr){4-5}
          & BiLSTM & CNN   & BiLSTM & CNN \\
    \midrule
    RND   & 0.688  & 0.658  & 0.531  & 0.594  \\
    EGL-Word  & 0.750  & 0.775  & 0.644  & 0.650  \\
    BALD  & 0.757  & 0.773  & 0.645  & 0.658  \\
    CORESET & 0.752  & 0.764  & 0.612  & 0.659  \\
    BADGE & 0.744  & 0.767  & 0.619  & 0.641  \\
    \themodel  & \textbf{0.803}  & \textbf{0.814}  & \textbf{0.700}  & \textbf{0.746}  \\
    \bottomrule
    \end{tabular}
  \label{tab:AUC}
\end{table}

\subsection{Baseline Approaches}

To evaluate the effectiveness of \themodel, we compare it with the following approaches:
\begin{itemize}
\item \textbf{RND} is a simple baseline which randomly selects samples in each iteration.
\item \textbf{EGL-Word} \cite{Zhang:2017Active} is an extension of EGL \cite{Huang:2016Active}, which utilizes norms of gradients to measure uncertainty for the task of text classification.
\item \textbf{BALD} \cite{Houlsby:2011Bayesian} is an uncertainty-based approach based on Bayesian inference. We apply dropout approximation \cite{Gal:2016Dropout,Gal:2017Deep} in our experiments, where the dropout rate is set to 0.5.
\item \textbf{CORESET} \cite{Sener:2018Active} uses the representations of the last layer in DNN as the representations.
\item \textbf{BADGE} \cite{Ash:2020Deep} can be viewed as a combination of EGL and CORESET.
\end{itemize}

\subsection{Experimental Settings}

To evaluate the performance of \themodel, we use two sentence classification datasets\footnote{\url{http://www.cs.cornell.edu/people/pabo/movie-review-data/}}: \textbf{Subj} \cite{Pang:2004Sentimental} and \textbf{MR} \cite{Pang:2005Seeing}, which contain 5000, 5331 positive samples and 5000, 5331 negative samples respectively.
In our experiments, we use accuracy as the evaluation metric.
We run each approach $10$ times and report the median of results.
We randomly select $60\%$, $20\%$, and $20\%$ samples in each dataset for training, validation, and testing respectively.
We train a word2vec\footnote{\url{https://code.google.com/archive/p/word2vec/}} model on each dataset to initialize the word embeddings and set the hidden dimensionality to $100$.
We use two deep models: BiLSTM and CNN for comprehensive evaluation.
For the implementation of BiLSTM, we use a single bidirectional LSTM layer with $100$ hidden units.
For the implementation of CNN, we set the filter size to $( 3,4,5)$ and set the hidden dimension to $100$ as well.
In both BiLSTM and CNN, we apply the ReLU activation and the dropout rate is set to $0.5$.
We use $2\%$ samples in the training set as the initial seed labeled set.
Furthermore, we label $2\%$ samples in the training set during each iteration until $50\%$ samples in the training set have been labeled.
In other words, we set $N$ yo $24$ and $K$ to $2\%$ of training samples for each dataset.

\subsection{Results and Analysis}

We present the learning curve of the performance with different ratios of labeled samples in Figure \ref{fig:curve}.
It is seen from the figure that in most cases, active learning approaches outperform random selection, which demonstrates the necessity of deep active learning.
EGL-Word and BALD perform similarly and they both slightly outperform CORESET and BADGE.
Meanwhile, it is clear that \themodel constantly outperform other compared approaches, which is demonstrated especially in the middle parts of the learning curves.

Additionally in Table \ref{tab:AUC}, we calculate the normalized area under curve scores of learning curves in Figure \ref{fig:curve}.
This metric evaluates the global performance of each compared approach and it is evident that \themodel achieves the best performance.
In summary, these results strongly demonstrate the advantages of our proposed \themodel approach.


\section{Conclusion}

In this paper, inspired by the local piece-wise interpretability of DNNs, we introduce the linearly separable regions of samples to the problem of deep active learning.
For the task of text classification, we propose a novel \themodel approach, which selects and labels samples according to the diverse interpretations of unlabeled sample.
Specifically, we use the most diverse interpretation of words in a sample to measure the sample diversity.
Experimental results on two text classification datasets with CNN and BiLSTM as classifiers show that the \themodel approach is able to consistently outperform state-of-the-art deep active learning approaches.

\begin{acks}
This work is supported by National Natural Science Foundation of China (U19B2038, 61772528) and Shandong Provincial Key Research and Development Program (2019JZZY010119).
\end{acks}

\balance
\bibliographystyle{ACM-Reference-Format}
\bibliography{ref}

\end{document}